\documentclass{article}
\usepackage{spconf,amsmath,graphicx}

\usepackage{floatrow}

\usepackage{epsfig}
\usepackage{makecell}
\usepackage{amsfonts}

\usepackage{xcolor}
\usepackage{subcaption}
\usepackage{multirow}
\usepackage{amssymb}

\usepackage{color, colortbl}
\definecolor{mygray}{gray}{0.9}

\let\OLDthebibliography\thebibliography
\renewcommand\thebibliography[1]{
  \OLDthebibliography{#1}
  \setlength{\parskip}{0pt}
  \setlength{\itemsep}{0pt plus 0.3ex}
}

\title{Query-based Video Summarization with Pseudo Label Supervision}
\name{Jia-Hong Huang$^{1}$$^{^\star}$, Luka Murn$^{2}$, Marta Mrak$^{2}$, Marcel Worring$^{1}$}
\address{
  $^{1}$University of Amsterdam, Amsterdam, Netherlands ;
  $^{2}$BBC Research and Development, London, UK}

\begin{document}\sloppy
%
\maketitle
\begin{abstract}
Existing datasets for manually labelled query-based video summarization are costly and thus small, limiting the performance of supervised deep video summarization models. Self-supervision can address the data sparsity challenge by using a pretext task and defining a method to 
acquire extra data with pseudo labels to pre-train a supervised deep model. In this work, we introduce segment-level pseudo labels from input videos to properly model both the relationship between a pretext task and a target task, and the implicit relationship between the pseudo label and the human-defined label. The pseudo labels are generated based on existing human-defined frame-level labels. To create more accurate query-dependent video summaries, a semantics booster is proposed to generate context-aware query representations. Furthermore, we propose mutual attention to help capture the interactive information between visual and textual modalities. Three commonly-used video summarization benchmarks are used to thoroughly validate the proposed approach. Experimental results show that the proposed video summarization algorithm achieves state-of-the-art performance.
\end{abstract}
\section{Introduction}
\label{sec:intro}

\begin{figure}[t!]
\begin{center}
\includegraphics[width=0.9\linewidth]{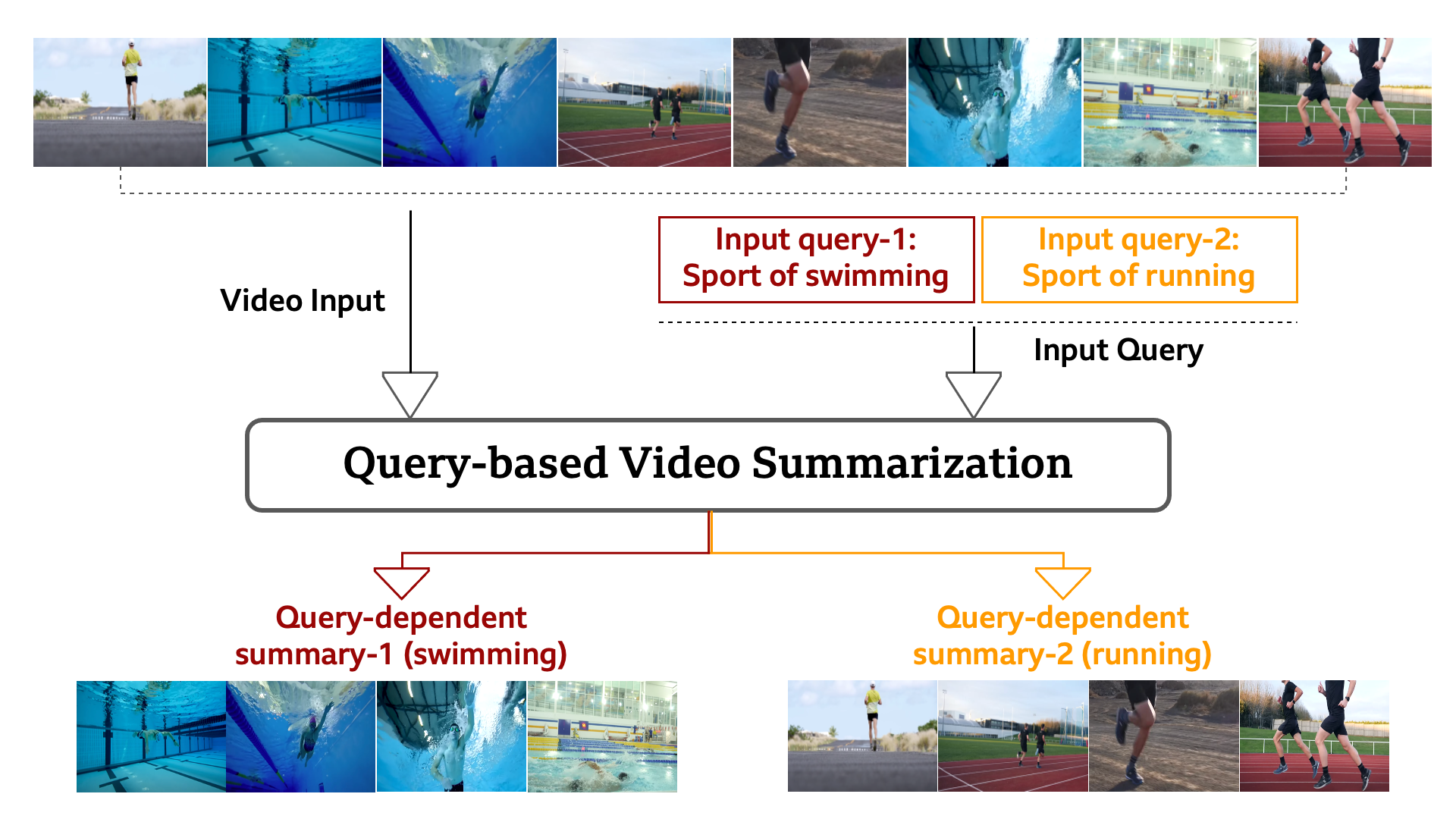}
\end{center}
\vspace{-0.2cm}
   \caption{Query-based video summarization. A video is summarized based on textual queries. The summarization algorithm runs independently for each query.}
\vspace{-0.5cm}
\label{fig:figure1}
\end{figure}

Query-based video summarization automatically generates a short video clip to summarize the content of a given video by capturing its query-dependent parts, as shown in Fig.~\ref{fig:figure1}. Such a  task can be modeled as a fully-supervised machine learning problem \cite{vasudevan2017query,huang2020query,huang2021gpt2mvs}. However, creating a large-scale manually-labeled video dataset for a fully-supervised task is costly. Hence, existing datasets, e.g., TVSum \cite{song2015tvsum}, SumMe \cite{gygli2014creating}, and QueryVS \cite{huang2020query}, are quite small.

\let\thefootnote\relax\footnotetext{$^{\star}$Work done during an internship at BBC Research and Development, London, UK.}

The lack of larger human-annotated datasets is common in fully-supervised deep learning tasks. Self-supervised learning is one of the most successful ways to alleviate this challenge \cite{doersch2015unsupervised,alwassel2019self,lai2019self,kim2019self}. According to \cite{alwassel2019self,caron2018deep}, self-supervision is an effective method to balance the cost of data labelling and the performance gain of a fully-supervised deep model. The main idea of self-supervised learning is defining a pretext task and introducing a way to acquire extra data with reliable pseudo labels to pre-train a fully-supervised deep model for performing a target task \cite{doersch2015unsupervised,alwassel2019self}.

Existing self-supervision methods assume that the relation between a target task with human-defined labels and an introduced pretext task with pseudo labels does not exist or exists in a very limited way \cite{alwassel2019self,caron2018deep}. However, this assumption may not be accurate for query-based video summarization, where frame-level human-defined labels can be considered as supervision signals of a target task. Segment-level pseudo labels can be considered as supervision signals of a pretext task. Since a video segment is composed of frames, there is an implicit relation between the entire segment and the corresponding frames. The improvement in model performance can hit a bottleneck without modelling these implicit relations.

In this work, a segment-based video summarization pretext task with specially designed pseudo labels is introduced to address this challenge, detailed in Fig.~\ref{fig:figure2}. Pseudo labels are generated based on existing human-defined annotations, helping to model the implicit relations between the pretext task and the target task, i.e., frame-based video summarization \cite{huang2020query,song2015tvsum,gygli2014creating}. In query-based video summarization, we observe that generating accurate query-dependent video summaries can be challenging in practice due to ineffective semantics embedding of textual queries. We address this issue by proposing a semantics booster that generates context-aware query representations which are capable of efficiently capturing the semantics. Furthermore, we noticed that the query input does not always help model performance, most likely due to the interactions between textual and visual modalities not being properly modelled. We address this challenge by introducing mutual attention that helps capture the interactive information between different modalities.

These novel design choices enable us to improve the model performance of query-based video summarization with self-supervision. Extensive experiments show that the proposed method is effective and achieves state-of-the-art performance. If we examine the problem from the perspective of frame-level label vs. segment-level label, the proposed method can also be considered as a weakly-supervised video summarization approach. Hence, existing weakly-supervised methods are also considered as baselines in this work.

\section{Related Work}

\subsection{Fully-supervised video summarization}
\vspace{-0.05cm}
Fully-supervised learning is a common way to model video summarization \cite{gygli2014creating,zhang2016video,zhao2017hierarchical,zhao2018hsa,jiang2022joint}. In fully-supervised video summarization, labels defined by human experts are used to supervise a model in the training phase. 
In \cite{gygli2014creating}, a video summarization approach is proposed to automatically summarize user videos that contain a set of interesting events. The authors start by dividing a video based on a superframe segmentation, tailored to raw videos. Then, various levels of features are used to predict the score of visual interestingness per superframe. Finally, a video summary is produced by selecting a set of superframes in an optimized way. In \cite{zhao2017hierarchical,zhao2018hsa}, a Recurrent Neural Network (RNN) is used in a hierarchical way to model the temporal structure in video data. The authors of \cite{zhang2016video} consider video summarization as a problem of structured prediction. A deep-learning-based method is proposed to estimate the importance of video frames based on modelling their temporal dependency. The authors of \cite{jiang2022joint} propose an importance propagation-based collaborative teaching network (iPTNet) for video summarization by transferring samples from a video moment localization correlated task equipped with a lot of training data. In \cite{huang2020query,huang2021gpt2mvs,huang2022causal,huang2023causalainer,yuan2017video,zhou2018video,wu2023expert,di2021dawn,yang2018novel,liu2018synthesizing,huck2018auto,huang2021longer,huang2021deep,huang2021contextualized,huang2022non,huang2021deepopht,huang2019assessing,huang2023improving,huang2017robustnessMS,huang2017robustness,huang2017vqabq,huang2019novel_1}, the model learning process expands beyond solely utilizing visual inputs and incorporates an additional modality, such as viewers' comments, video captions, or any other contextual data available. 

The aforementioned fully-supervised methods exploit a full set of human expert annotations to supervise the model in the training phase. Although such a method performs well, it is costly. Therefore, a better solution should be developed for video summarization.
\vspace{-0.45cm}
\subsection{Weakly-supervised video summarization}
\vspace{-0.05cm}
In \cite{panda2017weakly,ho2018summarizing,cai2018weakly,hu2019silco,chen2019weakly}, video summarization is considered as a weakly-supervised learning task. Weakly-supervised learning can mitigate the need for extensive datasets with human expert annotations. Instead of using a full set of data with human expert labels, such as frame-level annotations, weakly-supervised approaches exploit less-expensive weak labels, such as video-level annotations from human experts. Although weak labels are imperfect compared to a full set of human expert annotations, they still can be used to train video summarization models effectively. 
\vspace{-0.45cm}
\subsection{Self-supervision in video summarization}
\vspace{-0.05cm}
In \cite{yan2020self,jiang2019comprehensive}, image pretext tasks \cite{alwassel2019self} are extended to video for self-supervision in video summarization. In \cite{yan2020self}, the keyframes of a video are defined as those which are very different in their optical flow features and appearance from the rest of the frames of the video. The authors of \cite{jiang2019comprehensive} claim that a good video sequence encoder should have the ability to model the correct order of video segments. Segments are selected from a given video based on a fixed proportion before feeding it into a neural network. They are randomly shuffled and used to train the neural network and distinguish the odd-position segments to control the difficulty of the auxiliary self-supervision task. 

Existing work related to self-supervision in video summarization is very limited, and they do not focus on query-based video summarization. To the best of our knowledge, our proposed method is one of the pioneer works of self-supervision in query-based video summarization.
\vspace{-0.45cm}
\subsection{Word embedding methods}
\vspace{-0.05cm}
According to \cite{ethayarajh2019contextual}, static word embeddings and contextualized word representations are commonly used to encode textual data. Both of them are more effective than the Bag of Words (BoW) method. Skip-gram with negative sampling (SGNS) \cite{mikolov2013distributed} and GloVe \cite{pennington2014glove} are well-known models for generating static word embeddings. According to \cite{levy2014linguistic,levy2014neural}, these models learn word embeddings iteratively in practice. However, it has been proven that both of them implicitly factorize a word-context matrix containing a co-occurrence statistic. 

The authors of \cite{ethayarajh2019contextual} mention that in static word embeddings methods, all meanings of a polysemous word must share a single vector because a single representation for each word is created. Hence, the contextualized word representations method is more effective than static word embeddings because of its context-sensitive word representations. In \cite{devlin2018bert,peters2018deep,radford2019language}, the proposed neural language models are fine-tuned to create deep learning-based models for a wide range of downstream natural language processing tasks. 

In this work, a contextualized word representation-based method is used to encode the text-based input query.


\begin{figure*}[ht!]
\floatbox[{\capbeside\thisfloatsetup{capbesideposition={left,center},capbesidewidth=4cm}}]{figure}[\FBwidth]
{
\caption{Flowchart of the proposed self-supervision method for query-based video summarization. The model is pre-trained by the textual-spatial features from the Mutual Attention Mechanism and pseudo segment-level labels. The completely trained video summary generator exploits the fully-connected layer to produce a frame-level score vector for the given input video and outputs the final query-dependent video summary.}
\label{fig:figure2}
}
{\includegraphics[width=10cm]{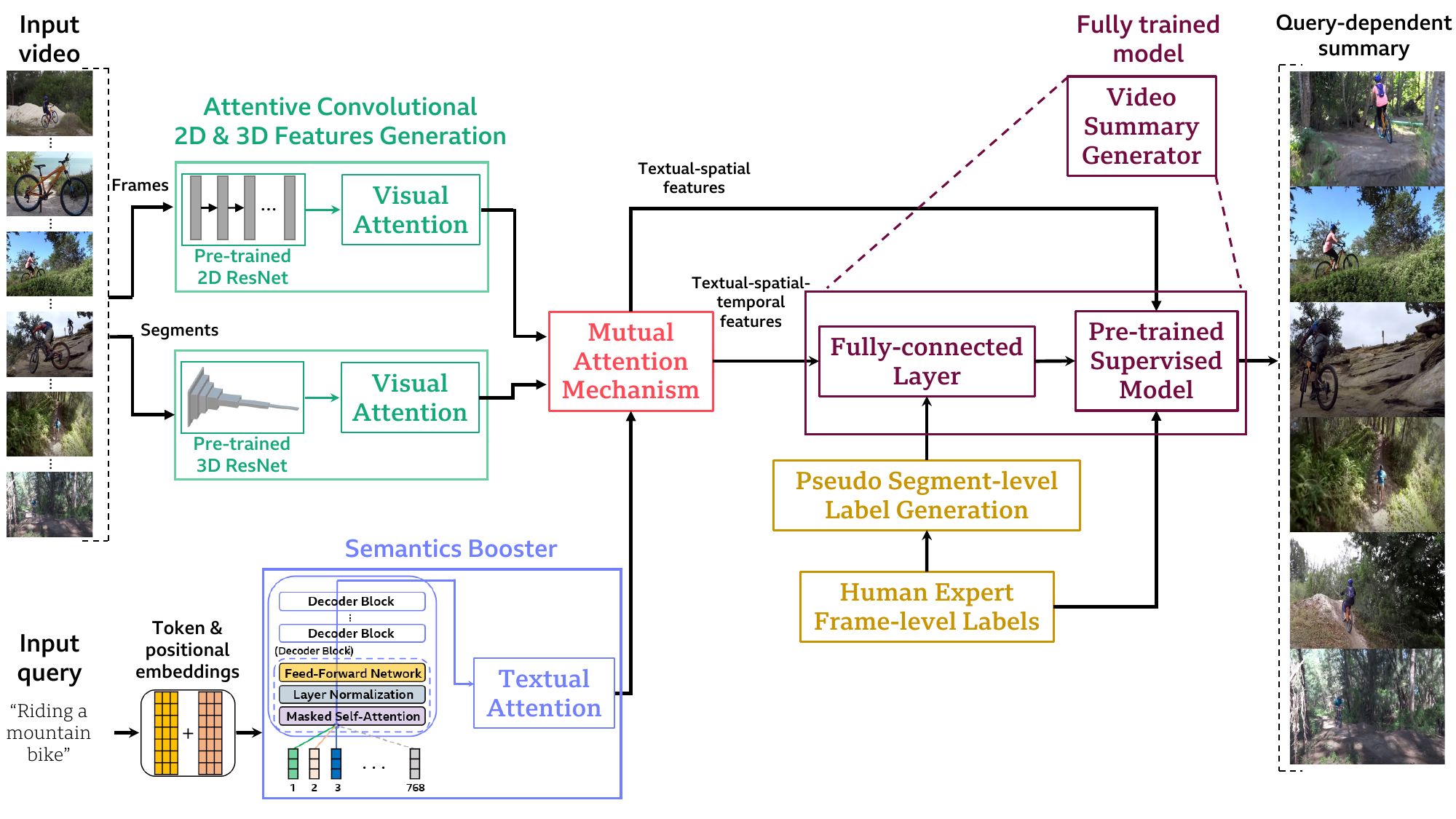}}
\vspace{-0.5cm}
\end{figure*}

\section{Methodology}

In this section, the proposed query-based video summarization 
method is described in detail, and illustrated in Fig.~\ref{fig:figure2}. The approach is based on contextualized query representations, attentive convolutional 2D and 3D features, interactive attention mechanism, mean-based pseudo shot label generation, and video summary generation.
\vspace{-0.45cm}
\subsection{Semantics Booster}
\vspace{-0.05cm}
Generating an accurate query-dependent video summary is challenging because of the ineffective semantics embedding of input textual queries. In this work, a semantics booster is introduced to capture the semantics of the input query effectively. The transformer-based model architecture has been firmly established as one of the state-of-the-art approaches in language modeling and machine translation \cite{vaswani2017attention}. Hence, the proposed semantics booster is built on top of the transformer architecture to generate context-aware query representations, described as follows.

For an input token $k_n$, its embedding $x_n$ is defined as: $x_n = W_e*k_n+P_{k_n}, n \in \{1,...,N\}$, where $W_e \in \mathbb{R}^{E_s \times V_s}$ is the input text-based query token embedding matrix with the vocabulary size $V_s$ and the word embedding size $E_s$, the positional encoding of $k_n$ is $P_{k_n}$, and $N$ denotes the number of input tokens. The subscripts $s$ and $e$ denote size and embedding, respectively. The representation of the current word $Q$ is generated by one linear layer defined as: $Q = W_q*x_n+b_q$, where $b_q$ and $W_q \in \mathbb{R}^{H_s \times E_s}$ are learnable parameters of the linear layer, the output size of the linear layer is $H_s$ and the subscript $q$ denotes query. The key vector $K$ is calculated by the other linear layer defined as: $K = W_k*x_n+b_k$, where $b_k$ and $W_k \in \mathbb{R}^{H_s \times E_s}$ are learnable parameters of the linear layer. The subscript $k$ denotes key. The value vector $V$ is generated by another linear layer defined as: $V = W_v*x_n+b_v$, where  $b_v$ and $W_v \in \mathbb{R}^{H_s \times E_s}$ are learnable parameters of the linear layer. The subscript $v$ denotes value. 

After $Q$, $K$, and $V$ are calculated, the masked self-attention is generated as: $\textup{MaskAtten}(Q,K,V) = \textup{softmax}(m(\frac{QK^T}{\sqrt{d_k}}))V$, where $m(\cdot)$ and $d_k$ denote a masked self-attention function and a scaling factor, respectively. The layer normalization is calculated as: $Z_{\textup{Norm}} = \textup{LayerNorm}(\textup{MaskAtten}(Q,K,V))$, where $\textup{LayerNorm}(\cdot)$ denotes a layer normalization function. Then, the introduced context-aware representation $\mathcal{R}_\textup{context}$ of the input text-based query is derived as: $\mathcal{R}_\textup{context} = \sigma(W_1Z_{\textup{Norm}}+b_1)W_2+b_2$, where  $\sigma$ is an activation function, $W_{1}$, $W_{2}$, $b_{1}$, and $b_{2}$ are learnable parameters of a position-wise feed-forward network.  To have even better textual representations, a textual attention function $\textup{TextAtten}(\cdot)$ is introduced to reinforce the context-aware representation. The function takes $\mathcal{R}_\textup{context}$ as input and calculates the attention and textual representation in an element-wise way. The attentive context-aware representation is calculated as $Z_{ta} = \textup{TextAtten}(\mathcal{R}_\textup{context})$, where $ta$ indicates textual attention.
\vspace{-0.45cm}
\subsection{Visual Attention}
\vspace{-0.05cm}
A 2D ConvNet and a 3D ConvNet are exploited to distill the video frame and video segment information, respectively. To reinforce the generated 2D and 3D features, a visual attention function $\textup{AttenVisual}(\cdot)$ is introduced to improve the quality of features. 

Let $E$ and $X$ be a feature generator and a set of video clips, respectively. A feature generator $E$ maps an input $x \in X$ to a feature vector $f \in \mathbb{R}^{d}$. $F=\{f=E(x) \in \mathbb{R}^{d} ~|~x \in X\}$ denotes a set of features produced by the feature generator $E$. Let $F_{s}$ be the generated features from the video spatial feature generator $E_{s}$. $F_{st}$ denotes the generated features from the video spatio-temporal feature generator $E_{st}$. Frame-level and segment-level data both are exploited to train the proposed query-based video summarization model, meaning $F = F_{s} \cup F_{st}$. In the frame-level case, the attentive feature generator $\textup{AttenVisual}(\cdot)$ learns attention weights and produces attentive spatial features $Z_{as}=\{f_{as}=\textup{AttenVisual}(f) \in \mathbb{R}^{d} ~|~f \in F_{s}\}$, i.e., attentive convolutional 2D features. In the segment-level case, the attentive feature generator learns attention weights and produces attentive spatio-temporal features $Z_{ast}=\{f_{ast}=\textup{AttenVisual}(f) \in \mathbb{R}^{d} ~|~f \in F_{st}\}$, i.e., attentive convolutional 3D features. 
\vspace{-0.4cm}
\subsection{Mutual Attention}
\vspace{-0.05cm}
We observe that textual queries do not always help the model performance due to the interactions between the video and query inputs not being modelled effectively. In this work, a mutual attention mechanism $\textup{MutualAtten}(\cdot)$ is introduced to address this issue and model the interactive information between the video and query. The mutual attention $Z_{ma}$ performs one by one convolution, i.e., convolutional attention. $Z_{ma} = \textup{MutualAtten}(Z_{ta} \odot Z_{as} \odot Z_{ast})$, where $Z_{ta}$ indicates textual attention and $\odot$ denotes Hadamard product.
\vspace{-0.45cm}
\subsection{Pseudo Segment-level Label Generation}
\vspace{-0.05cm}

Let $S_{f}$ be a set of human experts' frame-level score annotations and $P$  a pseudo score annotation generator that maps frame-level human expert scores to a segment-level pseudo score. 

In \cite{song2015tvsum}, the authors empirically find that a two-second segment is suitable for capturing local context of a video as it achieves good visual coherence. Based on this observation, in this work the proposed pseudo label generator $P$ is designed to generate a segment-level score every two seconds. In practice, since the generated pseudo score annotations are not validated by human experts, they might contain  noisy or biased information. Based on \cite{zhi2004analysis}, the $\textup{Mean}$ function is one of the effective ways to reduce the noise contained in the segment-level pseudo label. Hence, $\textup{Mean}$ function is used to design the proposed pseudo label generator $P$ to produce the mean score $S_{\textup{mean}}=P(S_{f})=\textup{Mean}(S_{f})$, i.e., the two-second segment-level pseudo score label. In the training phase, compared with the frame-level label, the mean-based pseudo segment label $S_{\textup{mean}}$ is used not only for spatial supervision but also for temporal supervision. The temporal supervision with the segment-level pseudo annotations improves the query-based video summarization model performance.
\vspace{-0.4cm}
\subsection{Loss Function}
\vspace{-0.05cm}
According to \cite{huang2020query}, query-based video summarization can be modeled as a classification problem. Thus, in this work, the categorical cross-entropy loss function is adopted to build the proposed approach:
\vspace{-0.2cm}
\begin{equation}
    \textup{Loss} = -\frac{1}{N}\sum_{i=1}^{N}\sum_{c=1}^{C}\mathbf{1}_{y_{i}\in C_{c}}\textup{log}(P_{\textup{model}}\left [y_{i}\in C_{c} \right ]),
    \label{eq:loss}
\vspace{-0.2cm}
\end{equation}
where $N$ indicates the number of observations, $C$ denotes the number of categories, $\mathbf{1}_{y_i \in C_c}$ is an indicator function of the $i$-th observation belonging to the $c$-th category, and $P_{\textup{model}}[y_i \in C_c]$ is the probability predicted by the model for the $i$-th observation to belong to the $c$-th category. 

\section{Experiments and Analysis}

\subsection{Datasets and evaluation metrics}
\vspace{-0.05cm}
\noindent \textbf{Datasets.} TVSum \cite{song2015tvsum} is a commonly used dataset for traditional video summarization, containing only the video input. However, authors of \cite{yuan2017video,zhou2018video} consider TVSum metadata, e.g., video title, as a text-based query input to generate the query-dependent video summary. In our experiments, the TVSum dataset is randomly divided into 40/5/5 videos for training/validation/testing, respectively. The video length is ranging from 2 to 10 minutes. The human expert score labels range from 1 to 5, and are annotated with 20 frame-level responses per video \cite{zhou2018video}. 

The SumMe \cite{gygli2014creating} dataset is randomly divided into 19 videos for training, 3 videos for validation, and 3 videos for testing. The video duration in SumMe is ranging from 1 to 6 minutes. In SumMe, the human expert annotation score ranges from 0 to 1. SumMe is not used for query-based video summarization and we do not have a query input when a model is evaluated on this dataset.

QueryVS \cite{huang2020query} is an existing dataset designed for query-based video summarization. In our experiments, the QueryVS dataset is separated into 114/38/38 videos for training/validation/testing, respectively. The video length in QueryVS is ranging from 2 to 3 minutes, and every video is retrieved based on a given text-based query.

To validate the proposed query-based video summarization method, three segment-level datasets are created based on the above frame-level datasets. Both the segment-level dataset, i.e., for pre-training, and the frame-level dataset, i.e., the target dataset, are used to conduct our experiments.

\noindent\textbf{Evaluation metric.} 
Based on \cite{song2015tvsum,gygli2014creating,yuan2017video,zhou2018video,hripcsak2005agreement}, the $F_{\beta}$-score with the hyper-parameter $\beta=1$ is a commonly used metric for assessing the performance of supervised video summarization approaches. It is based on measuring the agreement between the predicted score and ground truth score provided by the human expert. The $F_{\beta}$-score is defined as: $F_{\beta}=\frac{1}{N}\sum_{i=1}^{N}\frac{(1+\beta ^{2})\times p_{i}\times r_{i}}{(\beta ^{2}\times p_{i})+r_{i}}$, where $r_{i}$ indicates $i$-th recall, $p_{i}$ indicates $i$-th precision, $N$ indicates number of $(r_{i}, p_{i})$ pairs, ``$\times$'' denotes scalar product, and $\beta$ is used to balance the relative importance between recall and precision. 

\begin{table}[t!]
    \caption{Ablation study of the pseudo segment-level label pre-training, semantics booster, and mutual attention mechanism using $F_{1}$-score.}
\vspace{-0.4cm}
\centering
\scalebox{0.75}{
\renewcommand{\arraystretch}{1.3}
\begin{tabular}{|c|c|c|c|c|}
\hline
\makecell{Pseudo label\\pre-training} & \makecell{Mutual\\attention} & \makecell{Semantics\\booster} &
\textbf{TVSum} & \textbf{QueryVS} \\ \hline
- & - & - & 47.5 & 50.8 \\ \hline
\checkmark & - & - & 61.3 & 52.9 \\ \hline
- & \checkmark & - & 58.9 & 52.0 \\ \hline
- & - & \checkmark & 56.4 & 52.3 \\ \hline
\checkmark & \checkmark & \checkmark & \textbf{68.4} & \textbf{55.3} \\ \hline
\end{tabular}}
\vspace{-0.3cm}
\label{table:table1}
\end{table}

\begin{table}[t!]
    \caption{Comparison with state-of-the-art video summarization methods based on the $F_{1}$-score, best highlighted in bold. ‘-’ denotes unavailability from previous work.}
\vspace{-0.3cm}
\centering
\scalebox{0.75}{
\renewcommand{\arraystretch}{1.3}
\begin{tabular}{|c|c|c|c|c|}
\hline
\textbf{Model} & \textbf{Method} & \textbf{TVSum} & \textbf{SumMe}  &  \textbf{QueryVS} \\
\hline
vsLSTM \cite{zhang2016video} & \multirow{5}{*}{\shortstack{Fully\\supervised}} & 54.2 & 37.6 & - \\ 
\cline{1-1}\cline{3-5}
H-RNN \cite{zhao2017hierarchical} & & 57.7 & 41.1 & - \\
\cline{1-1}\cline{3-5}
HSA-RNN \cite{zhao2018hsa} & & 59.8 & 44.1 & - \\
\cline{1-1}\cline{3-5}
iPTNet \cite{jiang2022joint} & & 63.4 & 54.5 & - \\
\cline{1-1}\cline{3-5}
SMLD \cite{chu2019spatiotemporal} & & 61.0 & 47.6 & - \\
\cline{1-1}\cline{3-5}
SMN \cite{wang2019stacked} & & 64.5 & \textbf{58.3} & - \\
\hline
FPVSF \cite{ho2018summarizing} & \multirow{2}{*}{\shortstack{Weakly\\supervised}} & - & 41.9 & - \\ 
\cline{1-1}\cline{3-5}
WS-HRL \cite{chen2019weakly} & & 58.4 & 43.6 & - \\
\hline
DSSE \cite{yuan2017video} & \multirow{5}{*}{\shortstack{Query\\based}} & 57.0 & - & - \\
\cline{1-1}\cline{3-5}
DQSN \cite{zhou2018video} & & 58.6 & - & - \\
\cline{1-1}\cline{3-5}
QueryVS \cite{huang2020query} & & - & - & 41.4 \\
\cline{1-1}\cline{3-5}
GPT2MVS \cite{huang2021gpt2mvs} & & - & - & 54.8 \\
\cline{1-1}\cline{3-5}
Ours & & \textbf{68.4} & 52.4 & \textbf{55.3} \\
\hline
\end{tabular}}
\vspace{-0.3cm}
\label{table:table2}
\end{table}
\vspace{-0.4cm}
\subsection{Experimental settings}
\vspace{-0.05cm}
In the experiments, a 2D ResNet-34 network pre-trained on the ImageNet database \cite{deng2009imagenet} is adopted to generate frame-level features for each input video. The $512$ features are extracted from the visual layer one layer below the classification layer. A 3D ResNet-34 pre-trained on the Kinetics benchmark \cite{carreira2017quo} is used in the experiments to generate segment-level features for each input video. The features with $512$ dimensions are located in the visual layer which is right after the global average pooling layer. 

The video lengths in the SumMe, TVSum  and QueryVS datasets vary, with the maximum number of frames in a video being $388$ for SumMe, $199$ for QueryVS, and $647$ for TVSum. A frame-repeating preprocessing technique \cite{huang2020query} is followed to make all the videos in each dataset the same length. 

The input size of the CNN is $224$ by $224$ with RGB channels. Every channel is normalized by standard deviation $=(0.2737, 0.2631, 0.2601)$ and mean $=(0.4280, 0.4106, 0.3589)$. PyTorch is used for the implementation and to train models for $100$ epochs with $1e-7$ learning rate. The Adam optimizer is used, with hyper-parameters set as $\epsilon=1e-8$, $\beta_{1}=0.9$, and $\beta_{2}=0.999$.

\begin{figure}[t!]
\begin{center}
\includegraphics[width=0.85 \textwidth, height=0.13\textheight]{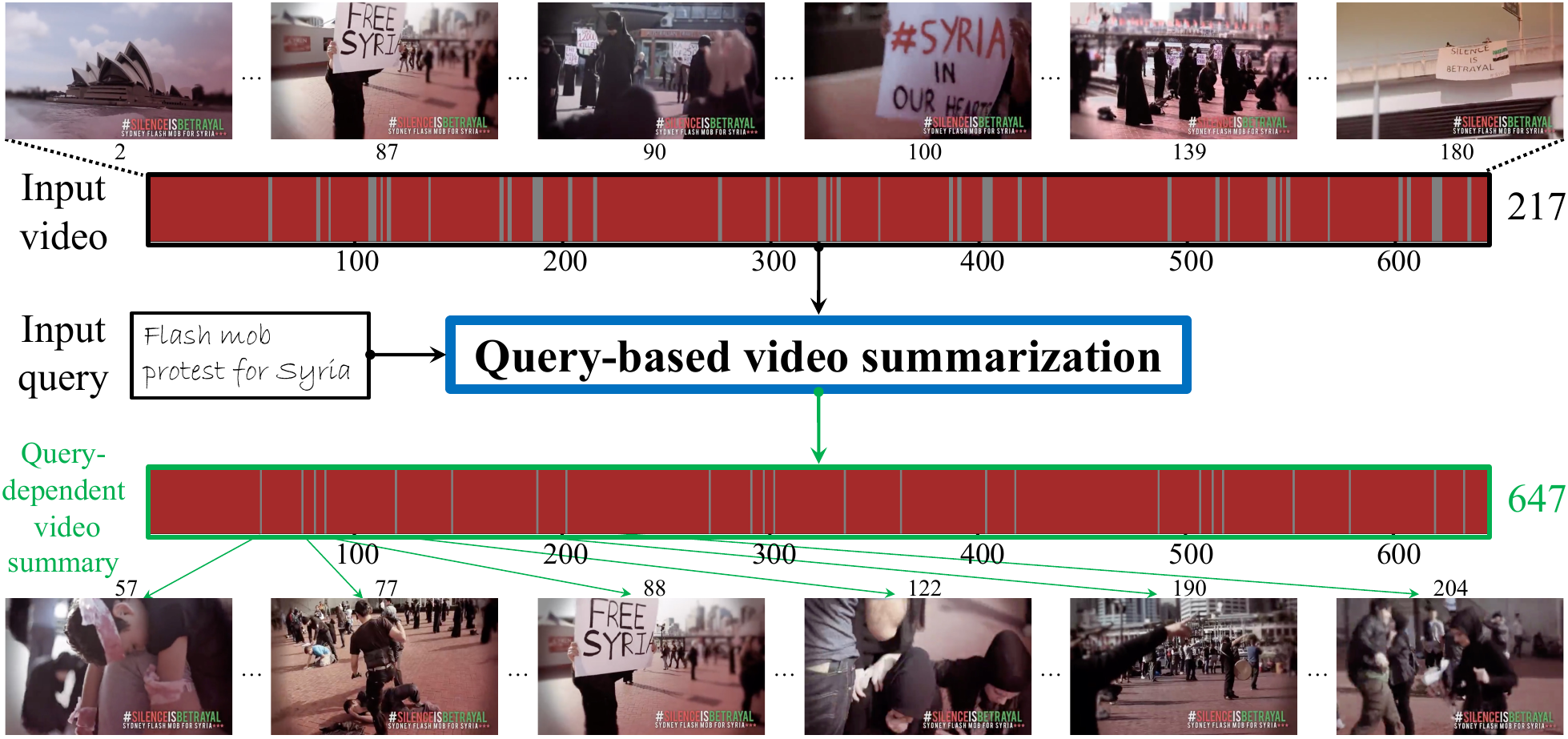}
\end{center}
\vspace{-0.1cm}
   \caption{Randomly selected qualitative results of the proposed method. Selected frames from the ground truth frame-based score annotations for the input video are highlighted in gray, with red representing the frames not selected. Frames selected for the query-dependent video summary are highlighted in green. $217$ denotes the video length before video preprocessing and $647$ denotes the video length after the video preprocessing.}
\vspace{-0.5cm}
\label{fig:figure3}
\end{figure}
\vspace{-0.4cm}
\subsection{Ablation Study}
\vspace{-0.05cm}
The ablation study of the proposed method is presented in Table~\ref{table:table1}. The baseline model without the mutual attention mechanism and pseudo segment label pre-training and no semantics booster performs significantly worse than approaches utilising any or all of the proposed improvements. Note that when the semantics booster is not adopted, the BoW embedding method is used.

The mutual attention mechanism helps capture the interaction between the input query and video more effectively. The pseudo segment-level label pre-training helps the proposed model have better initialization. The semantics booster captures the semantic meaning of the text-based query.
\vspace{-0.4cm}
\subsection{Comparison with state-of-the-art models}
\vspace{-0.05cm}
The comparison with existing fully-supervised, weakly-supervised and query-based approaches is presented in Table~\ref{table:table2}. The results show the performance of our proposed method is the best on TVSum and QueryVS datasets, with a competitive performance on the SumMe dataset. 

The correctness of the generated segment-level pseudo labels is not guaranteed by human experts, but it still contains useful information, e.g., better temporal information, to supervise the proposed model during pre-training. In weakly-supervised methods, although the correctness of the coarse labels, e.g., video-level label, is guaranteed by human experts, it is still not good enough to boost the model performance better than our proposed method. In query-based summarization methods, although the other modality is used to help the model performance, the effectiveness of the multi-modal feature fusion could limit the performance improvement.

Randomly selected qualitative results are shown in Fig.~\ref{fig:figure3}.

\section{Conclusion}

In this work, a new query-based video summarization approach is proposed. The method is based on the self-supervision of segment-level pseudo scores, semantics booster, and a mutual attention mechanism. Additionally, three segment-level video summarization datasets for self-supervision are proposed based on existing small-scale query-based video summarization datasets. Experimental results show the mean-based segment-level pseudo labels provide effective temporal supervision. The proposed approach achieves state-of-the-art performance in terms of the $F_1$-score. 
Nowadays, video content is growing at an ever-increasing speed and beyond the capacity of an individual for full comprehension. In such cases, the proposed query-based video summarization method has the potential to improve the efficiency of video exploration.

\section{Acknowledgments}
This project has received funding from the European Union’s Horizon 2020 research and innovation programme under the Marie Skłodowska-Curie grant agreement No 765140.

\bibliographystyle{IEEEbib}
\bibliography{strings,refs}

\end{document}